\documentclass{article}

\usepackage{arxiv}
\usepackage[utf8]{inputenc}
\usepackage[T1]{fontenc}
\usepackage{hyperref}
\usepackage{url}
\usepackage{hyperref}
\usepackage{booktabs}      
\usepackage{amsfonts}
\usepackage{nicefrac}
\usepackage{microtype}
\usepackage{lipsum}
\usepackage{graphicx}
\usepackage{amsmath}
\usepackage{algorithm}
\usepackage{algorithmic}
\usepackage{multirow}  
\usepackage{graphicx}
\usepackage{subcaption}
\usepackage{graphicx}
\usepackage{subcaption}
\usepackage{float}        
\usepackage[section]{placeins} 

\graphicspath{ {./images/} }

\title{Enhancing Reasoning Accuracy In Large Language Models During Inference Time}

\author{ \href{https://orcid.org/0000-0000-0000-0000}{\includegraphics[scale=0.06]{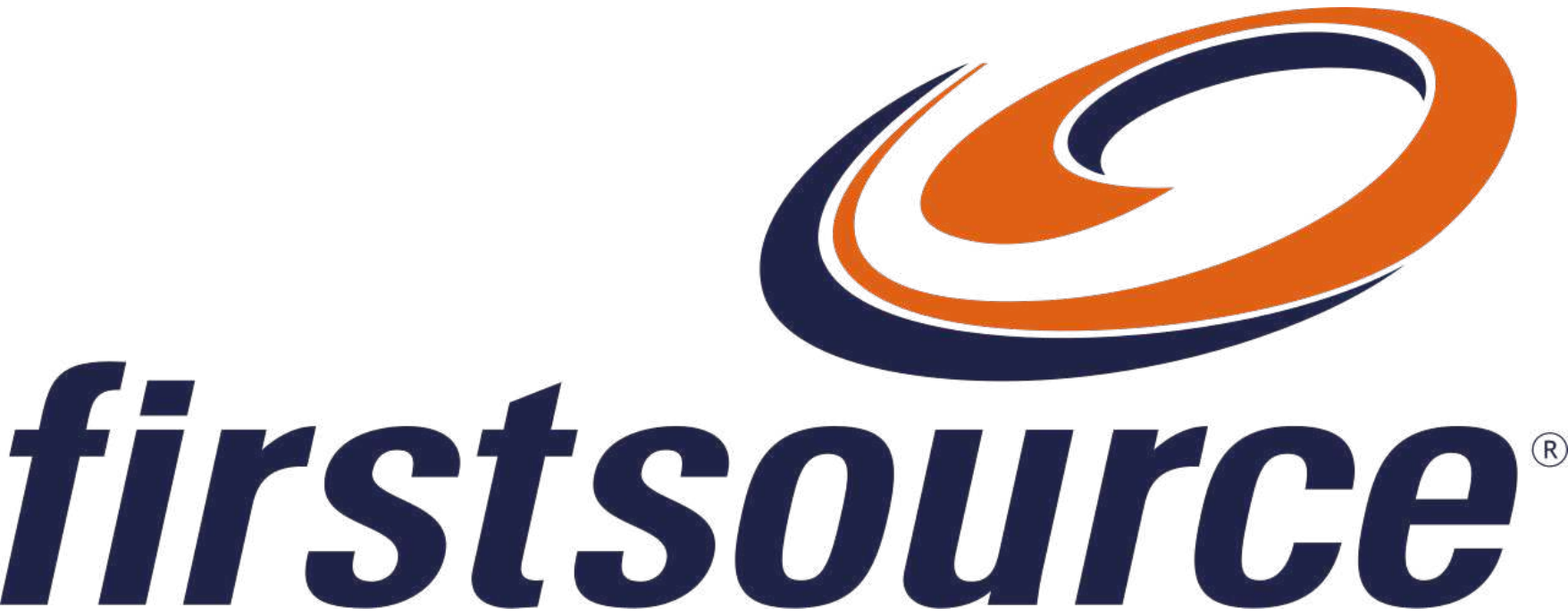}\hspace{1mm}Vinay Sharma \textsuperscript{*}}\\
	\texttt{Vinay.Sharma4@firstsource.com} \\
	\And
	\href{https://orcid.org/0000-0000-0000-0000}{\includegraphics[scale=0.06]{orcid.pdf}\hspace{1mm}Manish Jain \textsuperscript{*}} \\
	\texttt{Manish.Jain1@firstsource.com} \\
}

\begin{document}

\maketitle

\begingroup
\renewcommand\thefootnote{1}
\footnotetext{\textsuperscript{*}Equal contribution}
\endgroup

\pagestyle{plain}
\begin{abstract}
Large Language Models (LLMs) often exhibit strong linguistic abilities while remaining unreliable on multi-step reasoning tasks, particularly when deployed without additional training or fine-tuning. In this work, we study inference-time techniques to improve the reasoning accuracy of LLMs. We systematically evaluate three classes of inference-time strategies: (i) self-consistency via stochastic decoding, where the model is sampled multiple times using controlled temperature and nucleus sampling and the most frequent final answer is selected; (ii) dual-model reasoning agreement, where outputs from two independent models are compared and only consistent reasoning traces are trusted; and (iii) self-reflection, where the model critiques and revises its own reasoning. Across all evaluated methods, we employ Chain-of-Thought (CoT) \cite{wei2022chainofthought} prompting to elicit explicit intermediate reasoning steps before generating final answers. In this work, we provide a controlled comparative evaluation across three inference-time strategies under identical prompting and verification settings.

Our experiments on LLM \cite{jain2025mortgagellm} show that self-consistency with nucleus sampling and controlled temperature value yields the substantial gains, achieving a 9\%–15\% absolute improvement in accuracy over greedy single-pass decoding, well-suited for low-risk domains, offering meaningful gains with minimal overhead. The dual-model approach provides additional confirmation for model reasoning steps thus more appropriate for moderate-risk domains, where higher reliability justifies additional compute. Self-reflection offers only marginal improvements, suggesting limited effectiveness for smaller non-reasoning models at inference time.

\end{abstract}

\section{Introduction}
Large Language Models (LLMs) \cite{jain2025mortgagellm} have demonstrated impressive performance across a wide range of natural language tasks, including text generation, summarization, translation, question answering, etc. Their success is largely driven by large-scale pretraining on diverse corpora, enabling strong linguistic fluency and pattern recognition. However, despite these capabilities, LLMs often struggle with tasks that require reliable multi-step reasoning, logical consistency, or structured problem solving.

Improving reasoning performance in LLMs has therefore become an active area of research. Prior work has explored a variety of approaches, including architectural modifications, specialized training objectives, reinforcement learning \cite{li2017deepreinforcement}, and instruction tuning. While effective, such methods typically require access to model weights, additional training data, or substantial computational resources. This motivates the study of inference-time techniques that aim to improve reasoning reliability without modifying model parameters.

Inference-time strategies operate solely at the level of prompting, decoding, or output aggregation. Recent studies have shown that sampling-based methods, such as self-consistency \cite{zhou2025bridgingprob_selfconsistency}, can improve reasoning accuracy by exploiting diversity in model outputs. Other approaches leverage multiple models to cross-validate reasoning, or encourage models to critique and revise their own outputs through self-reflection \cite{liu2024selfreflectionllm}.

In this work, we present a systematic empirical study of inference-time reasoning techniques for improving the accuracy of LLMs on multi-step reasoning tasks.

\section{Dataset Details}

\textbf{Logical Reasoning Improvement Dataset:} All inference-time reasoning strategies were evaluated using the Logical Reasoning Improvement Dataset, publicly available on huggingface. The dataset was curated and released by garage-bAInd/Open-Platypus \cite{platypus2023}.

This dataset is particularly appropriate for our experimental setting for several reasons.

1. Dataset 'Input' text columns contains data from diverse domains like Science, Mathematics, Coding etc. thus allowing us to measure the effectiveness of inference-time techniques independently of specific domains.

2. The problems are deterministic and have unambiguous ground-truth answers, which enables reliable quantitative evaluation of accuracy improvements across decoding and verification strategies.

\textbf{Dataset Link:} https://huggingface.co/datasets/garage-bAInd/Open-Platypus

\textbf{Dataset Overview:} The dataset is organized into several columns, each serving a specific purpose:

input: The input text or question that requires logical reasoning. This column provides the initial statement or problem that needs solving.

output: The correct answer or solution to the logical reasoning question. This column contains the expected outcome or response.

instruction: Additional instructions or guidelines for solving the logical reasoning question. This column provides any specific guidance or steps required to arrive at the correct answer.

data source: The source or origin of the logical reasoning question. This column specifies where the question was obtained from.

\section{Methodology}

\subsection{Self Consistency + Controlled Temperature and Nucleus Sampling}

\textbf{Technique Overview:}
Self-consistency is an inference-time strategy designed to improve the reliability of Large Language Models (LLMs) on reasoning-intensive tasks without modifying model parameters or requiring additional training. Instead of relying on a single deterministic generation, the model is sampled multiple times under controlled stochastic decoding, and the final prediction is obtained by aggregating these independent reasoning paths.

In our implementation, self-consistency is combined with controlled temperature sampling and nucleus (top-p) sampling to encourage diverse yet semantically plausible reasoning trajectories. Specifically, the same question is posed to the model multiple times (n=6), each time allowing stochastic variation in token generation. Each run produces a full step-by-step reasoning trace followed by a final answer. These final answers are then extracted and then using a majority voting mechanism the most consistent answer across samples is determined.

This approach operationalizes the hypothesis that correct reasoning paths are more likely to recur across multiple stochastic generations than incorrect ones, even when individual generations are imperfect.

\textbf{Controlled Stochastic Decoding:}
To induce meaningful diversity across reasoning paths, we configure the reasoning model with a moderate temperature (T=0.8) and nucleus sampling (top p=0.9). 

Temperature controls the randomness of token selection by flattening the output distribution, while nucleus sampling restricts token choices to the smallest set whose cumulative probability mass exceeds p. Together, these mechanisms ensure that:

1. The model explores alternative reasoning trajectories rather than repeating a single deterministic path.

2. Generated responses remain coherent and grounded, avoiding degenerate or low-probability outputs.

\textbf{Answer Extraction and Majority Voting:}
Each reasoning output contains both intermediate steps and a final answer. To ensure fair aggregation, we employ a low-temperature model (T=0.1) to extract only the final answer from each generated response. This normalization step removes variability introduced by formatting, verbosity, or reasoning style.

Once extracted, the candidate answers are passed to a majority voting module, implemented as an LLM-based judge operating under near-deterministic settings. Rather than performing a simple string match, this judge selects the most semantically consistent answer across candidates. This design choice improves robustness to minor lexical variations (e.g., “2” vs. “two”) and aligns better with human notions of answer equivalence.

\textbf{Verification and Evaluation:}
For evaluation purposes, the final aggregated answer is compared against the ground-truth label using an independent verifier model.

\begin{figure}[H]
\centering
\includegraphics[width=\textwidth]{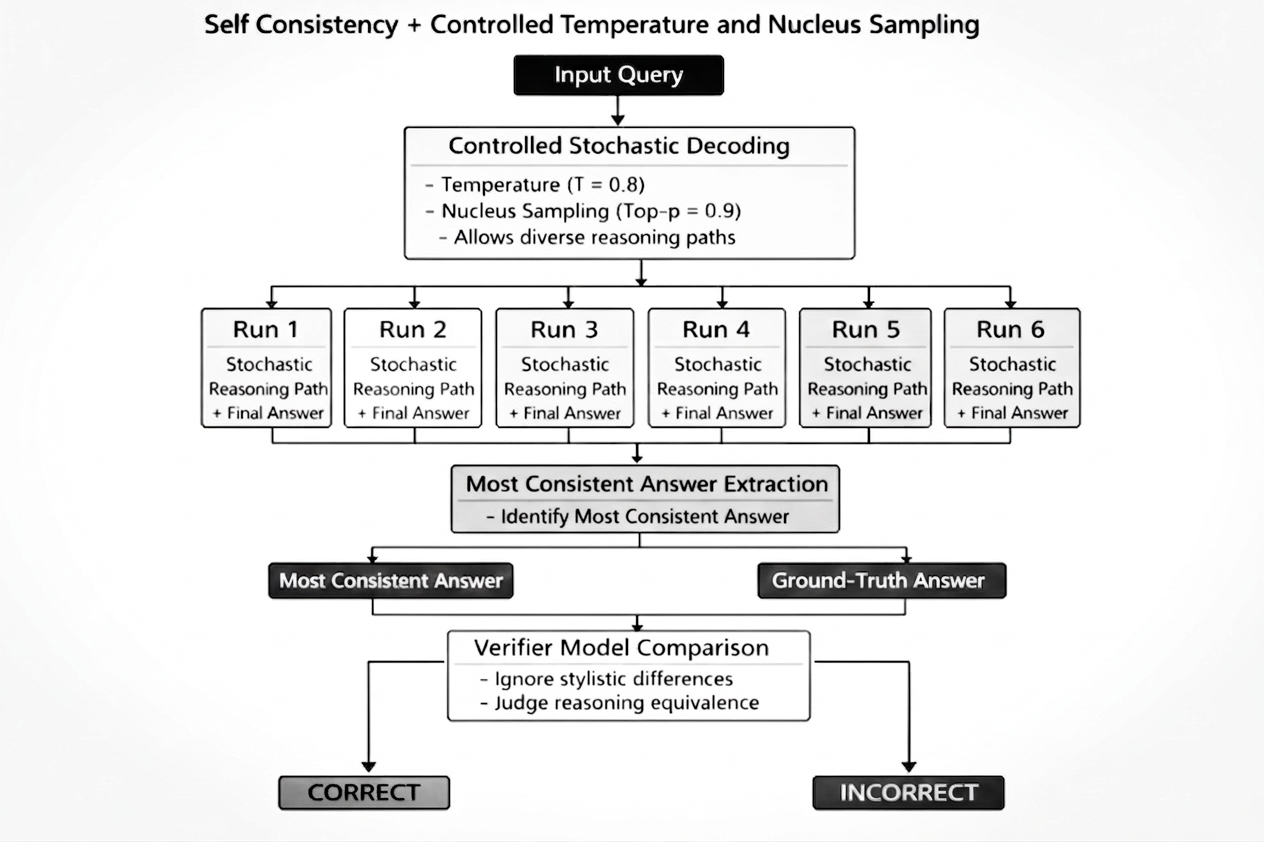}
\caption{Self Consistency With Controlled Temperature and Nucleus Sampling}
\end{figure}

\subsection{Dual-Model Reasoning with Cross-Model Verification}

\textbf{Technique Overview:}
Dual-Model Reasoning is an inference-time strategy that improves reasoning reliability by cross-validating solutions generated by two independent language models with different architectures, training data, and deployment characteristics. In our implementation, two models are prompted with the same reasoning instruction to independently solve the same problem. Each model produces a structured, step-by-step solution along with a final answer. A independent verifier model then evaluates whether both reasoning outputs reach the same correct conclusion. Agreement between models is treated as a strong signal of correctness, while disagreement flags the response as unreliable.
This approach is motivated by the observation that different models tend to fail in different ways, making cross-model agreement a powerful indicator of reasoning correctness.

\textbf{Cross-Model Agreement Verification:}
After both models generate their solutions, a verifier model is used to assess agreement. Rather than performing a superficial string comparison, the verifier is instructed to: 

Ignore differences in wording, formatting, and style.

Focus solely on whether both outputs arrive at the same correct conclusion.

Return a strict binary verdict: ACCEPT or REJECT.

\textbf{Verification and Evaluation:}
For evaluation, each model’s reasoning trace is independently checked against the ground-truth answer using a separate verification prompt. This allows us to measure not only cross-model agreement, but also whether agreement corresponds to factual correctness.

\textbf{Trade-offs and Practical Considerations:}
While Dual-Model Reasoning improves reliability, it introduces higher inference cost and latency compared to single-model approaches. As a result, this technique is best suited for scenarios where accuracy and trustworthiness are prioritized over throughput, or where model agreement is used as a gating mechanism before downstream actions or human review.

\begin{figure}[H]
\centering
\includegraphics[width=\textwidth]{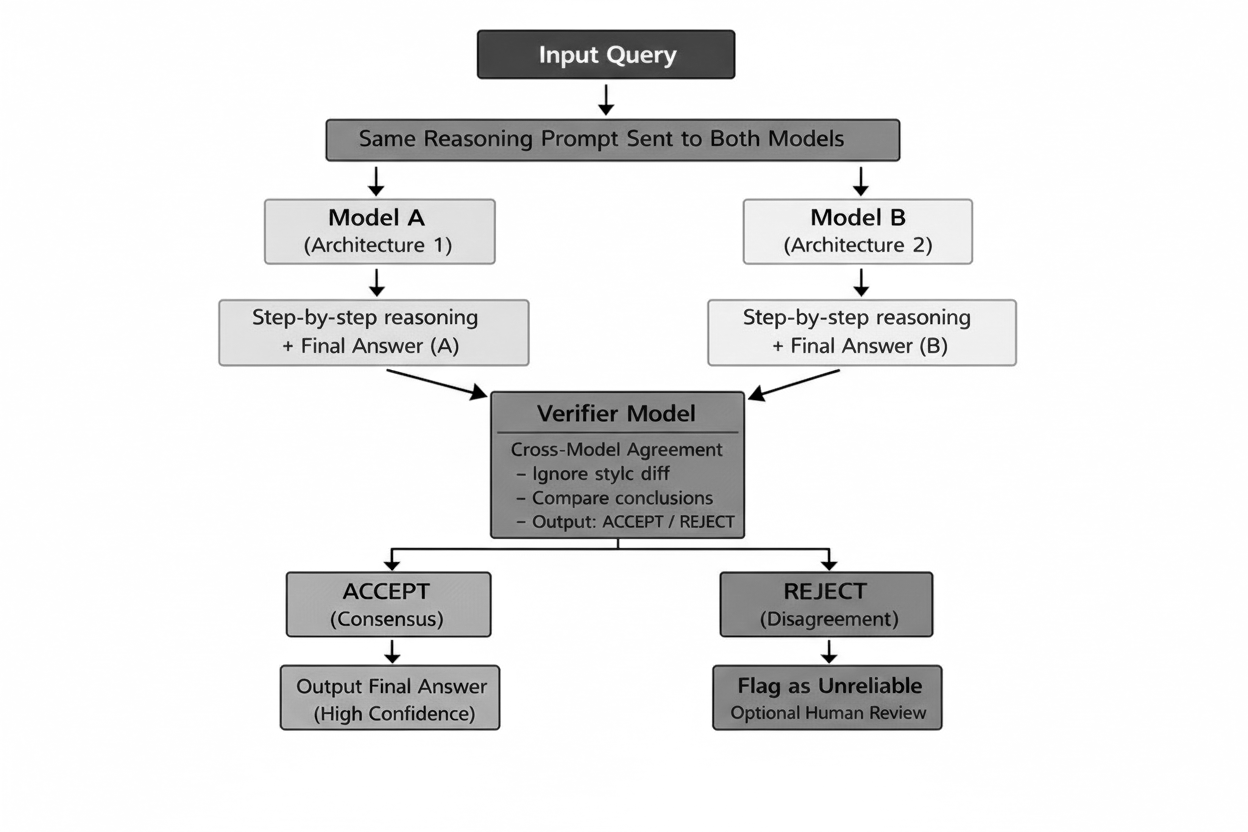}
\caption{Cross Model Verification Flow}
\end{figure}

\subsection{Self-Reflection via Iterative Critique and Revision}

\textbf{Technique Overview:}
Self-Reflection is an inference-time reasoning strategy in which a single language model critiques and revises its own solution before producing a final answer. 

Rather than relying on multiple samples or multiple models, this technique treats reasoning as an iterative refinement process, where errors are identified and corrected through self-evaluation.

In our implementation, the model first generates an initial step-by-step solution to a given question. This solution is then passed back to the same model under a reflection prompt that explicitly instructs it to analyze the reasoning, identify logical flaws, missing steps, or incorrect assumptions, and to explicitly acknowledge correctness when no errors are found. 

Finally, a revision step incorporates this critique to produce an improved solution and final answer.

\textbf{Verification and Evaluation:}
To evaluate the effectiveness of self-reflection, both the initial reasoning output and the revised output are independently verified against the ground-truth answer using a separate verifier model. 

This setup enables a direct comparison between pre-reflection and post-reflection accuracy, isolating the contribution of the reflection step itself.

\begin{figure}[H]
\centering
\includegraphics[width=0.8\textwidth]{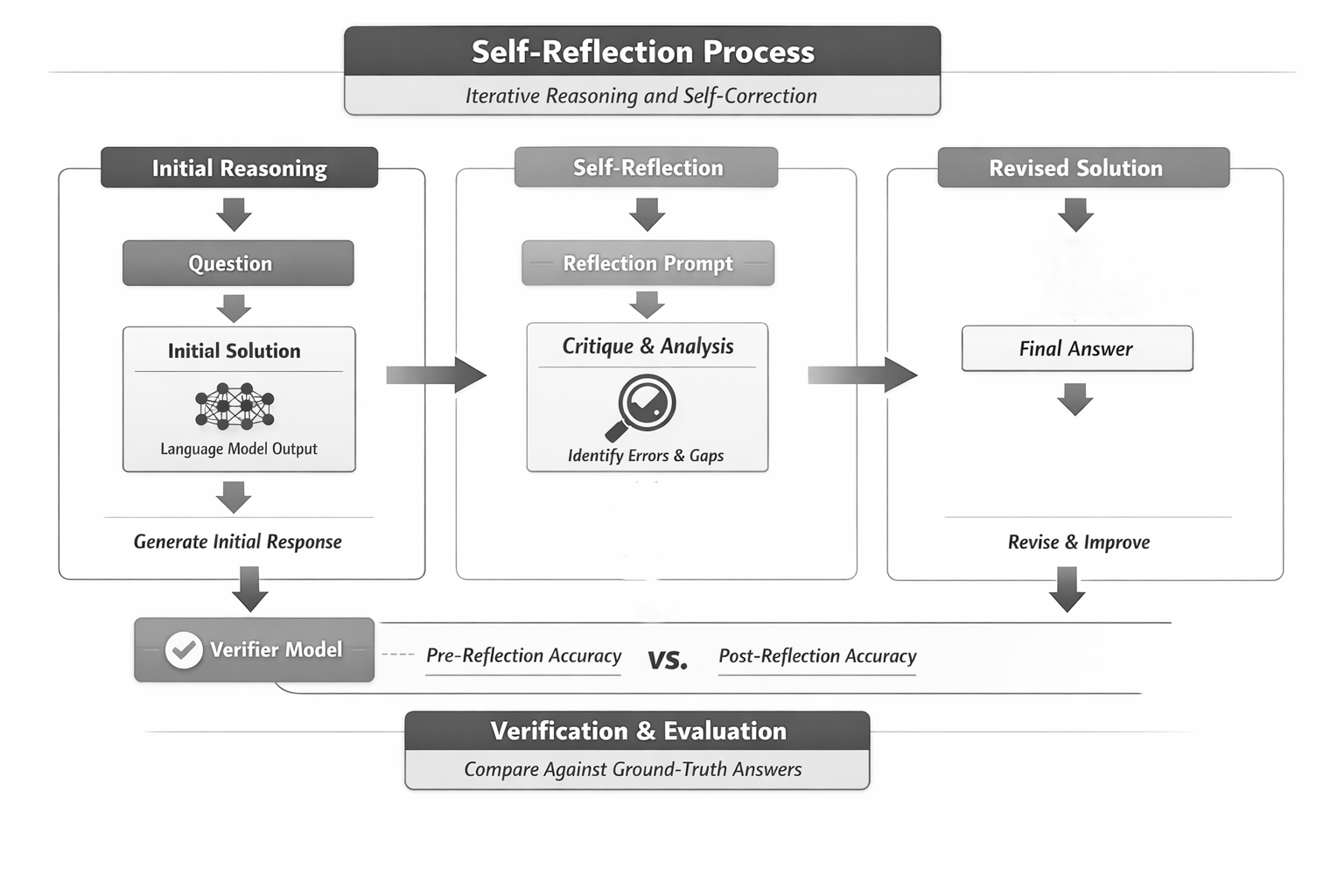}
\caption{Self-Reflection via Iterative Critique and Revision}
\end{figure}

\section{Observations}

\subsection{Self Consistency + Controlled Temperature and Nucleus Sampling}

Our experiments reveal a clear performance gap between self-consistency implemented with controlled temperature and nucleus sampling and self-consistency combined with greedy decoding.

When self-consistency is applied with controlled stochastic decoding (temperature = 0.8 and top p=0.9), the model achieves a substantially higher acceptance rate. Specifically, 64.9\% of model outputs are verified as correct, while 35.1\% are rejected.

In contrast, self-consistency combined with greedy decoding exhibits a noticeably lower acceptance rate. Under greedy decoding, only 56.2\% of outputs are accepted, with 43.8\% rejected.

Overall, these findings suggest that self-consistency implemented with controlled temperature and nucleus sampling, plays a critical role in improving reasoning accuracy and should be considered a necessary component of self-consistency–based approaches.

\begin{figure}[H]
\centering
\includegraphics[width=0.5\textwidth]{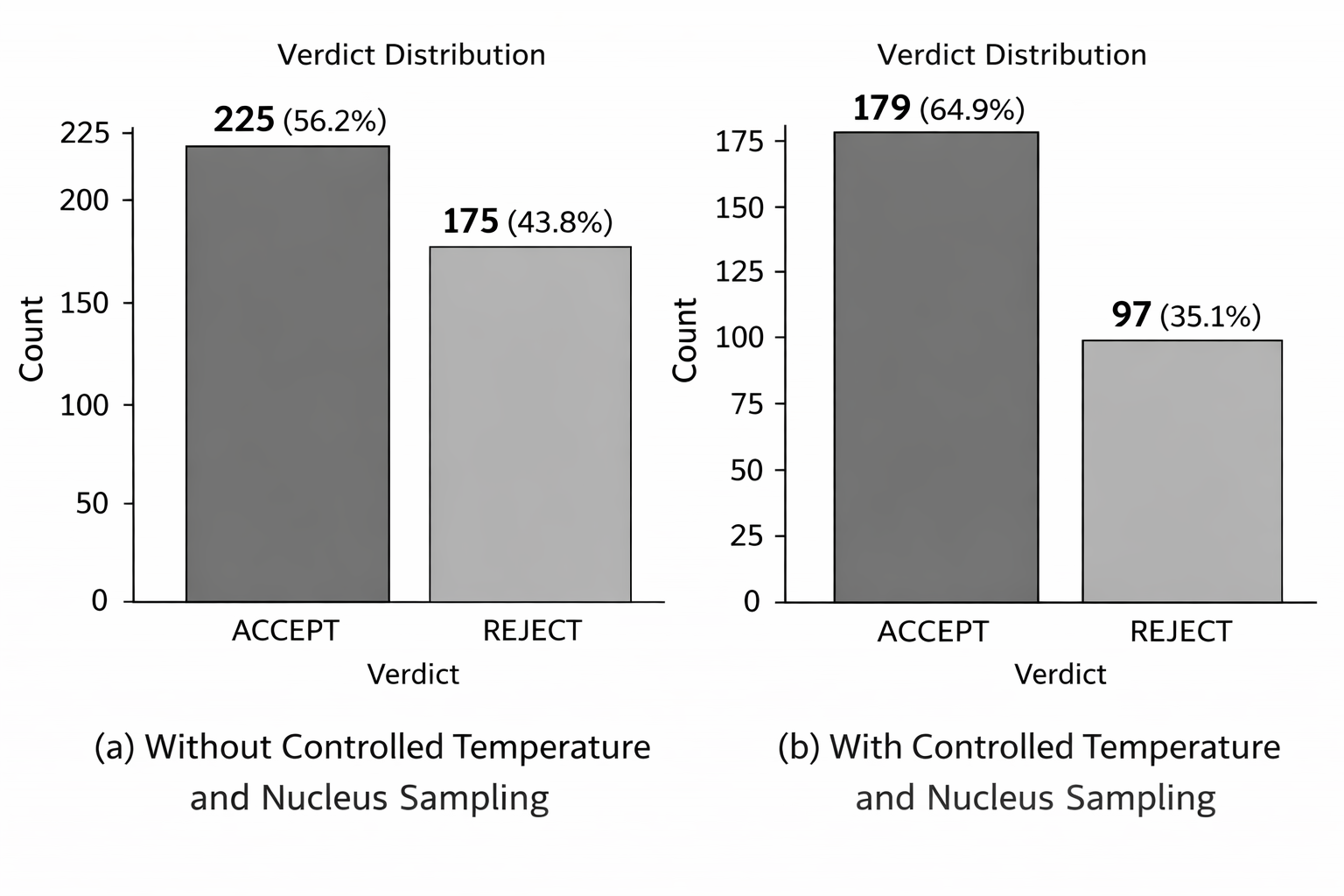}
\caption{Verdict Distribution Comparison for Self-Consistency Decoding}
\end{figure}

\subsection{Dual-Model Reasoning with Cross-Model Verification}

The results for Dual-Model Reasoning highlight both the limitations of single-model reasoning and the practical value of cross-model agreement in the absence of ground truth.

When evaluated against ground-truth answers, the original single-model responses achieve an acceptance rate of 48.7\%, indicating that fewer than half of the generated solutions are correct. After applying Dual-Model Reasoning with cross-model verification, the acceptance rate slightly decreases to 47.4\%. While this may appear counterintuitive at first glance but the post-verification acceptance rate closely aligns with the ground-truth–validated correctness. This observation is particularly significant for real-world deployment scenarios where ground-truth answers are unavailable or expensive to obtain. In such settings, Dual-Model Reasoning provides a principled mechanism for estimating and filtering reliable outputs based solely on inference-time signals. By accepting only those responses that survive cross-model scrutiny, the system prioritizes precision over recall, reducing the risk of propagating incorrect reasoning.

Overall, these results demonstrate that Dual-Model Reasoning is best viewed not as an accuracy-boosting technique in labeled evaluation settings, but as a confidence-estimation and validation mechanism. It is especially valuable in high-risk or open-ended environments, where correctness cannot be directly verified and conservative acceptance criteria are desirable.

\begin{figure}[H]
\centering
\includegraphics[width=0.6\textwidth]{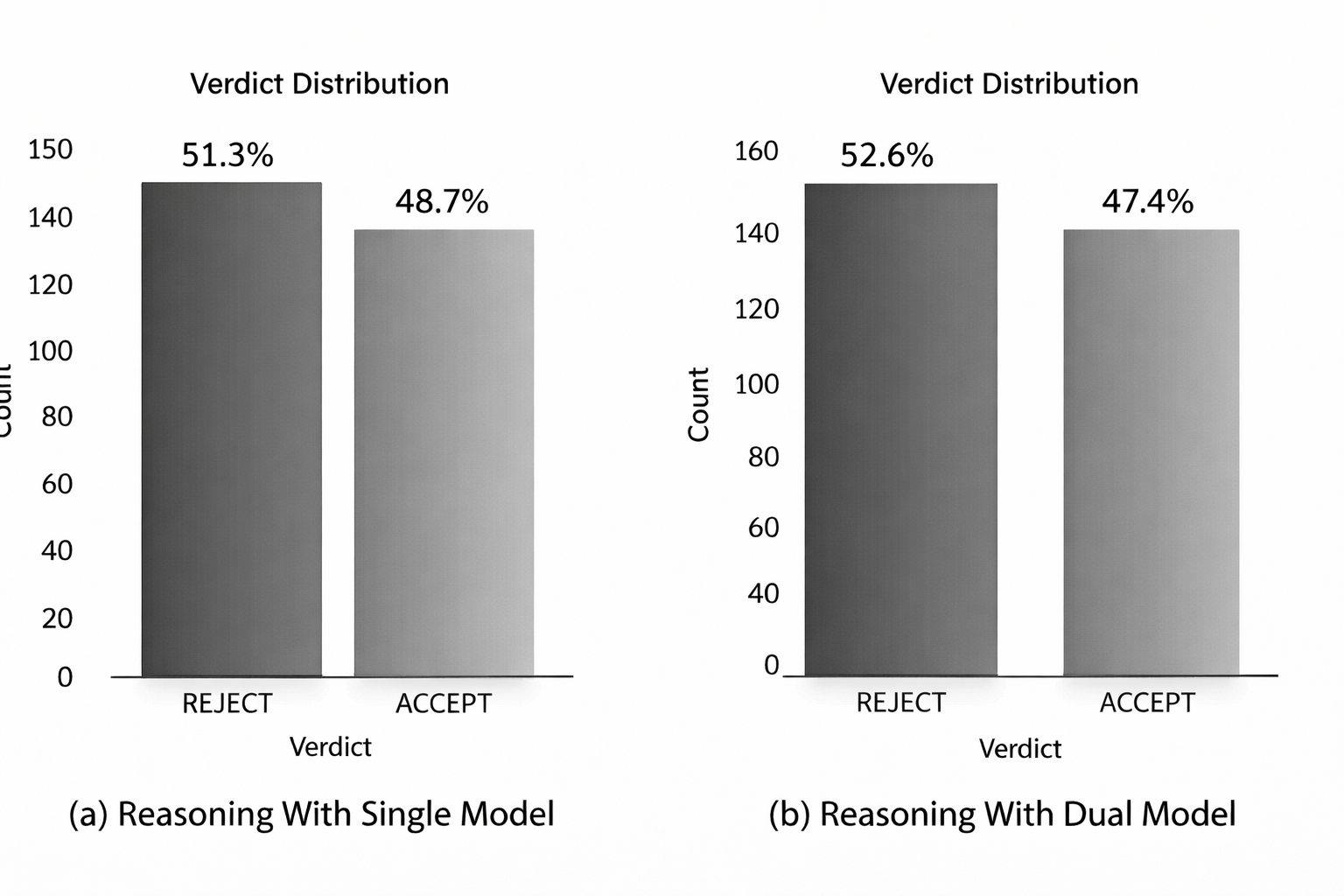}
\caption{Verdict Distribution Comparison for Dual-Model Reasoning}
\end{figure}

\subsection{Self-Reflection via Iterative Critique and Revision}
The results for Self-Reflection indicate that only marginal improvements are achieved when the technique is applied to a non-reasoning smaller model.

In the baseline setting, the model’s initial step-by-step reasoning achieves an acceptance rate of 47.2\%, with 52.8\% of responses rejected. This confirms that single-pass inference using a smaller, non-specialized reasoning model struggles on multi-step reasoning tasks. After introducing the self-reflection pipeline—consisting of explicit critique followed by revision—the acceptance rate increases modestly to 50.6\%, while the rejection rate decreases to 49.4\%. These observations indicate that self-reflection provides limited benefit when applied in isolation to smaller models.

\begin{figure}[h!]
\centering
\includegraphics[width=0.6\textwidth]{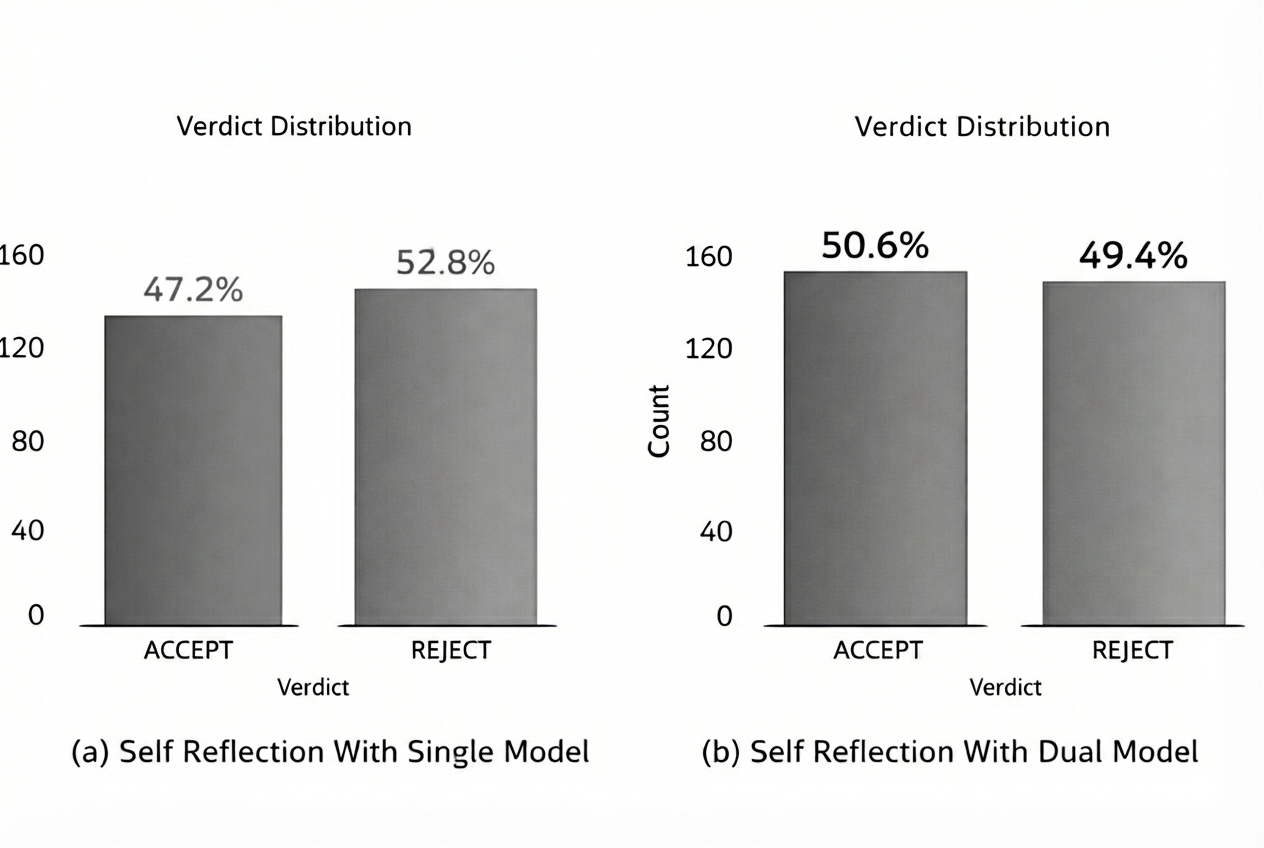}
\caption{Verdict Distribution Comparison for Self-Reflection Reasoning}
\end{figure}

\section{Conclusion}
In this work, we systematically evaluated three inference-time strategies for improving reasoning accuracy in Large Language Models without modifying model parameters: (i) self-consistency with controlled temperature and nucleus sampling, (ii) dual-model reasoning with cross-model verification, and (iii) self-reflection via iterative critique and revision. Our results demonstrate that inference-time techniques can significantly bolster the reasoning capabilities of Large Language Models (LLMs) without the need for resource-intensive retraining. The choice of strategy should be closely aligned with the risk profile and operational requirements of the specific domain.

For domains like finance, healthcare, legal, compliance support where the cost of a reasoning error can be catastrophic, Dual-Model Reasoning is the most appropriate strategy. This approach prioritizes precision over recall, acting as a critical gating mechanism that flags unreliable outputs for human review. In domains such as education, customer support assistance, or general information retrieval, Self-Consistency via Stochastic Decoding offers the best balance of performance and efficiency. While Self-Reflection showed only marginal gains (approximately 3.4 percentage points) for smaller models, it suggests that iterative refinement may require more specialized or larger architectures to be truly effective.

In summary, inference-time strategies provide a flexible toolkit for improving LLM reliability. By selecting the right technique—ranging from efficiency-focused self-consistency to safety-focused dual-model verification—organizations can deploy LLMs more confidently across diverse sectors.

\bibliographystyle{unsrt}  
\bibliography{references}

\end{document}